\documentclass[runningheads]{llncs}
\usepackage{graphicx}
\usepackage{epsfig}

\usepackage{amsmath}
\usepackage{amssymb}
\usepackage{graphicx} 
\usepackage{times} 
\usepackage{bm}
\usepackage{subfig}
\usepackage{url}
\usepackage{multirow}
\usepackage{float}
\usepackage{textcomp}
\usepackage{xcolor}
\usepackage{makecell} 

\begin{document}
	\title{Bayesian dense inverse searching algorithm for real-time stereo matching in minimally invasive surgery}
	\titlerunning{Bayesian dense inverse searching for stereo matching}
	%
	\author{Jingwei Song\inst{1,2} \and%
Qiuchen Zhu\inst{3} \and%
Jianyu Lin\inst{4} \and%
Maani Ghaffari\inst{1}}%
\authorrunning{J. Song et al.}
%
\institute{Michigan Robotics, University of Michigan, Ann Arbor, MI 48109, USA
\email{\{jingweso,maanigj\}@umich.edu}\\
\and
United Imaging Research Institute of Intelligent Imaging, Beijing 100144, China\\
\and
School of Electrical and Data Engineering, University of Technology, NSW 2007, Australia\\
\email{Qiuchen.Zhu@uts.edu.au}
\and
Hamlyn Centre for Robotic Surgery, Imperial College London, London SW7 2AZ, UK\\
\email{xjtuljy@gmail.com}
}
	\maketitle              
	\begin{abstract}
		This paper reports a CPU-level real-time stereo matching method for surgical images ($10$ Hz on $640 \times 480$ image with a single core of i5-9400). The proposed method is built on the fast LK algorithm, which estimates the disparity of the stereo images patch-wisely and in a coarse-to-fine manner. We propose a Bayesian framework to evaluate the probability of the optimized patch disparity at different scales. Moreover, we introduce a spatial Gaussian mixed probability distribution to address the pixel-wise probability within the patch. In-vivo and synthetic experiments show that our method can handle ambiguities resulted from the textureless surfaces and the photometric inconsistency caused by the non-Lambertian reflectance. Our Bayesian method correctly balances the probability of the patch for stereo images at different scales. Experiments indicate that the estimated depth has similar accuracy and fewer outliers than the baseline methods in the surgical scenario with real-time performance. The code and data set are available at \url{https://github.com/JingweiSong/BDIS.git}.\par
		
	\end{abstract}
	%
	%
	%
	
	\section{Introduction}
	
	Real-time 3D intra-operative tissue surface shape recovery from stereo images is important in Computer Assisted Surgery (CAS). The reconstructed depth is a crucial for dense Simultaneous Localization and Mapping (SLAM) \cite{song2017dynamic,song2018mis}, AR system \cite{haouchine2013image,widya20193d} and diseases diagnosis \cite{mahmood2019polyp,jia2020automatic}. All stereo matching procedures follow the pinhole camera model \cite{andrew2001multiple} and conduct image rectification, undistortion, and disparity estimation. The stereo matching techniques are normally classified into two categories regarding disparity estimation: prior-free and learning-based. Conventional prior-free methods estimate the pixel-wise disparity using the image alignment techniques \cite{hirschmuller2005accurate,stoyanov2010real,geiger2010efficient,kroeger2016fast,rappel2016surgical}. Based on the left-right image consistency assumption (photo-metric or feature-metric), they either use corner feature registration, dense direct pixel searching, or a combination. Differently, Deep Neural Network (DNN) based techniques directly learn the disparity from the training image pairs \cite{ye2017self,turan2018deepvo,yang2019hierarchical,brandao2020hapnet,long2021dssr}. Although DNN methods are reported to be efficient, the results may be invalidated with changing parameters such as focal length and baseline or a large texture difference between the training and testing data \cite{pratt2014practical,allan2021stereo}. Moreover, the DNN-based methods heavily depend on the size and quality of the annotated training data, which are not accessible in many CAS scenarios. \par 
	
	In the category of prior-free methods, ELAS~\cite{geiger2010efficient} is still one of the most widely used stereo matching algorithms due to its robustness and accuracy \cite{song2017dynamic,song2018mis,zhang2017autonomous,zhan2020autonomous}. It is also the most popular method in the industry \cite{zampokas2018real,cartucho2020visionblender}. ELAS uses Sobel descriptors to match sparse corners as the supporting points and triangulate the pixel-wise disparity prior. Then, the optimal dense disparity is retrieved with its proposed maximum a-posteriori algorithm. Its two-step process requires around $0.25-1$ second on a single modern CPU core. This paper aims for a faster CPU-based stereo matching method.\par
	
	The Dense Inverse Searching (DIS) \cite{kroeger2016fast} shows the potential of dense direct matching without the time-consuming sparse supporting points alignment. By resizing the left and right images to several coarse scales, it adopts and modifies the Lucas-Kanade (LK) optical flow algorithm \cite{lucas1981iterative} for fast estimating the pixel-wise optimal disparity. \cite{kroeger2016fast} demonstrates that real-time computation is possible with its patch-based coarse-to-fine dense matching, where patch refers to an arbitrary squared image segment. However, DIS is strictly built based on the photometric consistency and surface texture abundance assumptions, which cannot always be satisfied in CAS. The two main challenges are the textureless/dark surfaces and the serious non-Lambertian reflectance. The weak/dark texture, which widely exists in CAS, leads to ambiguous photometric consistency. Meanwhile, non-Lambertian reflectance brings uneven disturbance on the surfaces, and it cannot be eliminated by just enforcing the patch normalization~\cite{shimasaki2013generating}.\par  
	
	In this paper, to deal with photometric inconsistency and non-Lambertian reflectance in stereo matching, we propose a Bayesian Dense Inverse Searching (BDIS) to quantify the posterior probability of each optimized patch. A spatial Gaussian Mixture Model (GMM) is further adapted to quantify pixel-wise confidence within the patch. The final pixel-wise disparity is the fusion of multiple local overlapping patches, reducing the impact of those patches suffering from the textureless/dark surfaces or the non-Lambertian reflectance. In extreme cases, it is beneficial to give up the disparity estimation of some patches identified as dubious. In particular, this work has the following contributions:
	
	\begin{itemize}
		\item A Bayesian approach is developed to quantify the posterior probability of the patch. 
		\item A spatial GMM is introduced to quantify the pixels' confidence within the patch. 
		\item To our knowledge, BDIS is the first single core CPU based stereo matching approach that achieves similar performance to the near real-time method ELAS.
	\end{itemize}
	
	\section{Methodology}
	
	\subsection{Multiscale DIS}
	\label{section_2_1}

\begin{figure}[!h]
		\centering
		\subfloat{
			\begin{minipage}[]{1\textwidth}
				\centering
				\includegraphics[width=1\linewidth]{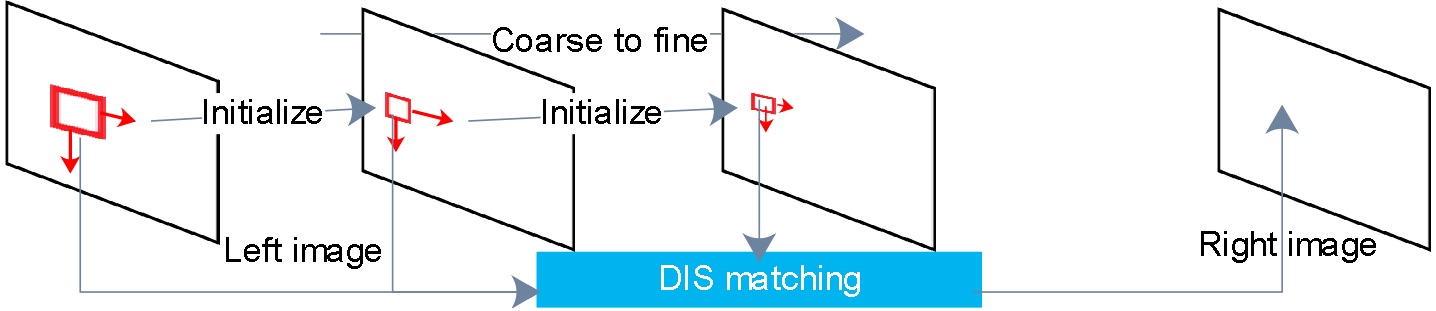}
			\end{minipage}
		}
		\caption{The framework of the DIS algorithm \cite{kroeger2016fast}. It uses 3 scale levels as an example.}
		\label{fig_framework}
	\end{figure}

	Fig. \ref{fig_framework} shows the DIS (based on fast LK) algorithm for stereo matching proposed by \cite{kroeger2016fast}. It is a modified version of the LK algorithm. We use the fast DIS as our base framework. Note that the variational refinement module in \cite{kroeger2016fast} is abandoned because it has a small (less than $0.5\%$) contribution in promoting the accuracy. The modified fast LK based DIS is achieved by minimizing the following objective function: 
	
	\begin{equation}
	\label{Eq_fast_inverse_search}
	\Delta \mathbf{u}=\operatorname{argmin}_{\Delta \mathbf{u}^{\prime}} \sum_{x}\left[I_{r}\left(\mathbf{x}+\mathbf{u}\right)-I_{l}(\mathbf{x}+\Delta \mathbf{u}^{\prime})\right]^{2},
	\end{equation}
	
	\noindent where $\mathbf{x}$ is the processed location, $\mathbf{u}$ is the estimated disparity in the loop, $I_l$ and $I_r$ are the left image patch and right image, and $\Delta \mathbf{u}$ is the optimal update of $\mathbf{u}$ at one loop. Different from authentic LK, $\Delta \mathbf{u}^{\prime}$ is moved from the right image to the left image patch. The improvement avoids the expensive re-evaluation of the Hessian on the right image. \eqref{Eq_fast_inverse_search} is traversed on all patches at different scales. The disparity at the fine-scale level is initialized at the optimized coarse scale. The optimal disparity at the location $\mathbf{x}$ is the weighted fusion with all covering patches using inverse residual:
	
	\begin{equation}
	\label{Eq_residual_fusion}
	\hat{\mathbf{u}}_\mathbf{x} = \sum_{k \in \Omega} \frac{ 1/ \operatorname{max}(\lVert I_l(\mathbf{x}+\mathbf{u}^{(k)})-I_r(\mathbf{x})\lVert^2,1)}{\sum_{k \in \Omega} 1/ \operatorname{max}( \lVert I_l(\mathbf{x}+\mathbf{u}^{(k)})-I_r(\mathbf{x})\lVert^2,1)}  \mathbf{u}^{(k)},
	\end{equation}
	\noindent where $\Omega$ is the set of patches covering the position $\mathbf{x}$, $\mathbf{u}^{(k)}$ is the estimated disparity of the patch $k$ and $\operatorname{max}(\cdot,\cdot)$ selectes the maximum value. The pixel-wise disparity $\hat{\mathbf{u}}_\mathbf{x}$ is the weighted average of the estimated disparities from all patches, wherein the weight is the inverse residual of brightness.\par
	
	\subsection{The Bayesian patch-wise posterior probability}
	\label{section_2_2}
	The residual-based weighted average fusion \eqref{Eq_residual_fusion} suffers from the ambiguities brought by the textureless/dark surface and non-Lambertian reflectance. The textureless/dark surface leads to ambiguous local minima of the cost function penalizing photometric inconsistency \eqref{Eq_fast_inverse_search} and misleads the algorithm to be over-confident on the estimation. Furthermore, the photometric consistency presumption is seriously violated on the surface affected heavily by the non-Lambertian reflectance. The affine lighting changes formulation in previous large-scale SLAM studies \cite{engel2017direct} cannot fully tackle the complex and severe non-Lambertian reflectance in CAS. In both situations, the weights retrieved from the photometric residuals \eqref{Eq_residual_fusion} are misleading. To overcome the difficulty in defining the confidence of the estimated disparity, we propose a Bayesian model to correctly estimate the confidence in the presence of textureless surface and non-Lambertian reflectance. Since the uncertainty distribution of both the left and right scenes is unclear, it is difficult to conduct the direct inference of the posterior probability in terms of disparity. Thus, we implicitly infer the probability with Bayesian modeling using Conditional Random Fields (CRF) \cite{uzunbas2016efficient}. The posterior probability of the patch-wise disparity $\mathbf{u}^{(k)}$ is
	
	\begin{equation}
	\begin{aligned}
	\label{Eq_post_probability}
	p(\mathbf{u}^{(k)}|I_l,I_r)\propto\frac{p(I_r|I_l,\mathbf{u}^{(k)})}{p(I_r,I_l,\mathbf{u}^{(k)})}\propto
	\frac{p(I_r|I_l,\mathbf{u}^{(k)})}{\Sigma_{\mathbf{u}_i^{(k)} \in \mathcal{P}} p(I_r|I_l,\mathbf{u}_i^{(k)})}\propto\frac{p(I_r|I_l,\mathbf{u}^{(k)})}{\Sigma_{\mathbf{u}_i^{(k)} \in \mathcal{P}'} p(I_r|I_l,\mathbf{u}_i^{(k)})}\mathrm{r},
	\end{aligned}
	\end{equation}
	
	\noindent where $\mathcal{P}$ is the domain of all possible choice of $\mathbf{u}_i^{(k)}$. To reduce computational load, $\mathrm{r}$ is applied as the constant compensation ratio for all patches within the window. $\mathcal{P}$ is reduced to a small window $\mathcal{P}'$ assuming the rest candidates are numerically trivial.\par 
	
	Equation \eqref{Eq_post_probability} indicates that the posterior probability of the disparity can be obtained by traversing the probability on all possible $\mathbf{u}_i^{(k)}$. And the possible choice of disparity is equal to window size $\mathrm{s}$. Even though the posterior probability suffers from the textureless surface and non-Lambertian reflectance, the illumination consistency probability is proportional to the residuals because the set of neighboring disparities is within one patch, and the impact of the issues is consistent. Thus, we model the illumination consistency probability $p(I_r|I_l,\mathbf{u}_i^{(k)},\mathbf{x})$ based on the Boltzmann distribution \cite{larochelle2008classification} as
	
	\begin{equation}
	\label{Eq_prior_probability}
	p(I_r|I_l,\mathbf{u}_i^{(k)}) = \exp\left(-\frac{ \lVert I_l(\mathbf{u}_i^{(k)})-I_r(\mathbf{u}_i^{(k)})\lVert^2_\mathrm{F}}{2\sigma_r^2\mathrm{s}^2}\right),
	\end{equation}  
	
	\noindent where $\lVert\cdot\lVert_\mathrm{F}$ is the Frobenius norm and $\sigma_r$ is the hyperparameter to describe the variance of the brightness. The relative posterior probability can be obtained with \eqref{Eq_post_probability} and \eqref{Eq_prior_probability}. Generally, the absolute exponential parameter of the Boltzmann distribution denotes the entropy of the state. In our case, such entropy is defined as \eqref{Eq_prior_probability}. Image with abundant texture has more entropy loss. Hence, the entropy item is highly related to the photometric inconsistency loss.  \par

\begin{figure}[!h]
		\centering
		\subfloat{
			\begin{minipage}[]{0.7\textwidth}
				\centering
				\includegraphics[width=1\linewidth]{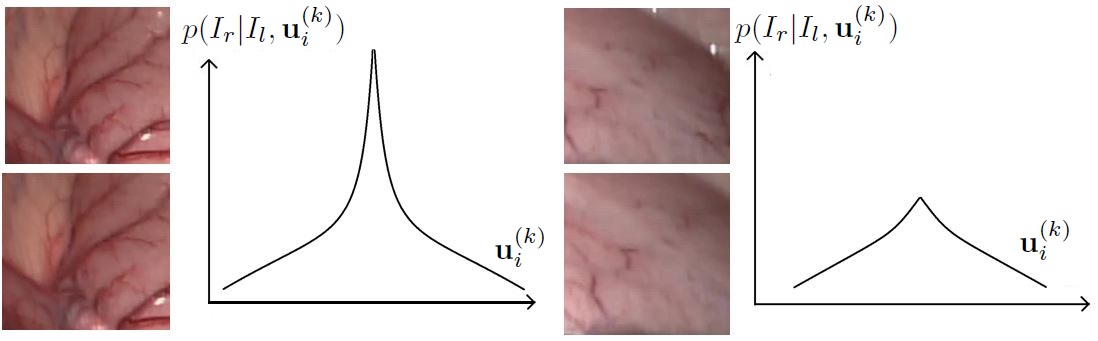}
			\end{minipage}
		}
		\caption{The probability density function of the textureless region.}
		\label{fig_prob_density}
	\end{figure}

	Fig. \ref{fig_prob_density} shows the relationship between the illumination consistency probability density function and the texture. The response is stronger on the textured surface. The residuals are always small in the textureless surface, no matter how the left and right images are aligned. \eqref{Eq_residual_fusion} cannot correctly measure the weights while \eqref{Eq_prior_probability} describes the relative probability of the estimation. Moreover, it tests the local convergence to filter the Saddle point solutions. \par

	\subsection{The prior spatial Gaussian probability}
	\label{section_2_3}
	
	In addition to the patch-wise posterior probability of the disparity in the last section, a spatial GMM is adopted to estimate pixel-wise probability within the patch. Considering that medical images are natural images, a multivariate Gaussian distribution is adopted to measure the confidence of the pixel-wise probability using a Gaussian mask. In accordance with the multivariate Gaussian distribution, the center of the patch has higher confidence than the edge pixels since those central pixels preserve more information for inference. Assuming all pixels in the patch are i.i.d, we have

	\begin{equation}
	\begin{aligned}
	\label{Eq_spatial_prob}
	p(\mathbf{u}^{(k)}|I_l,I_r,\mathbf{x}) \propto p(\mathbf{u}^{(k)}|I_l,I_r)\exp\left(-\frac{\sum_{\mathbf{\xi}^{(k)}(\mathbf{x})}\lVert\mathbf{x}-\mathbf{\xi}^{(k)}(\mathbf{x})\lVert^2_\mathrm{F}}{2\sigma_s^2}\right),
	\end{aligned}
	\end{equation}

	\noindent where $\mathbf{\xi}^{(k)}(\mathbf{x})$ is the set of all pixel positions within the patch $k$ in image coordinate. $\sigma_s$ is the 2D spatial variance of the probability. Note that \eqref{Eq_spatial_prob} is independent of the patch and can therefore be pre-computed before the process. Combining \eqref{Eq_post_probability}, \eqref{Eq_prior_probability} and \eqref{Eq_spatial_prob}, the final pixel-wise posterior probability distribution can be represented as follows,
	
	\begin{equation*}
	\begin{aligned}
	\label{Eq_final_prob}
	p(\mathbf{u}^{(k)}\!|\!I_l,I_r,\mathbf{x})\! \propto\! \exp\left(\!-\frac{\!\sum_{\!\mathbf{\xi}^{(k)}(\mathbf{x})}\!\lVert\mathbf{x}-\mathbf{\xi}^{(k)}(\mathbf{x})\lVert^2_\mathrm{F}}{2\sigma_s^2}\right) \frac{\!\exp\left(-\frac{\!\lVert I_l(\mathbf{u}^{(k)})-I_r(\mathbf{u}^{(k)})\!\lVert^2_\mathrm{F}}{2\sigma_r^2\mathrm{s}^2}\!\right)}
	{\!\sum_{\mathbf{u}_i^{(k)}\! \in\! \mathcal{P}}\exp\left(-\frac{ \lVert I_l(\mathbf{u}_i^{(k)})-I_r(\mathbf{u}_i^{(k)})\lVert^2_\mathrm{F}}{2\sigma_r^2\mathrm{s}^2}\!\right)}.
	\end{aligned}
	\end{equation*}  
	
	Finally, it should be emphasized that \eqref{Eq_prior_probability} and \eqref{Eq_spatial_prob} are not the cost functions but probability/weight for each patch or pixel. Costly optimization steps are avoided.\par 
	
	
	
	%
	%
	%
	
	\section{Results and discussion}
	
	BDIS was compared with DIS \cite{kroeger2016fast}, SGBM \cite{hirschmuller2005accurate} and ELAS \cite{geiger2010efficient} on the in-vivo and the synthetic data sets \footnote{Readers are encouraged to watch the attached video and test the code.}. The computations were implemented on a commercial desktop (i5-9400) in C++. DNN-based methods PSMNet \cite{chang2018pyramid} and GwcNet \cite{guo2019group} were also compared for completeness and the computation was conducted on the GTX 1080ti in PyTorch. The public in-vivo stereo videos from \cite{giannarou2013probabilistic} were adopted which contains 200 images with size $640\times480$ and 200 images with size $288\times 360$. All stereo images were rectified, undistorted, calibrated, and vertically aligned with the provided intrinsic and extrinsic parameters. We also provided a synthetic data set generated from an off-the-shelf virtual phantom of a male's digestive system. A virtual handheld colonoscope was placed inside the colon and was manipulated to go through the colon to collect the depth and stereo images. The 3D game engine Unity3D\footnote{\url{https://unity.com/}} was used to generate the sequential stereo and depth images with a pin-hole camera in size $640 \times 480$. The synthetic distortion-free data has accurate intrinsic and extrinsic parameters. Both diffuse lighting ($100$ frames) and non-Lambertian reflectance ($100$ frames) were simulated. $\gamma$ was set to $0.75$ for $640\times480$ and $0.25$ for $288\times 360$ data to discard the patch without enough valid pixels. $\sigma_r$ and $\sigma_s$ were set to $4$; the sampling within one Bayesian window was 5; the disturbance from the convergence was $0.5$ and $1$ pixel.\par
	
	\begin{figure}[htpb]    
	\centering
	\subfloat{    
		\begin{minipage}[htpb]{1\textwidth}    
			\centering
			\includegraphics[width=1\linewidth]{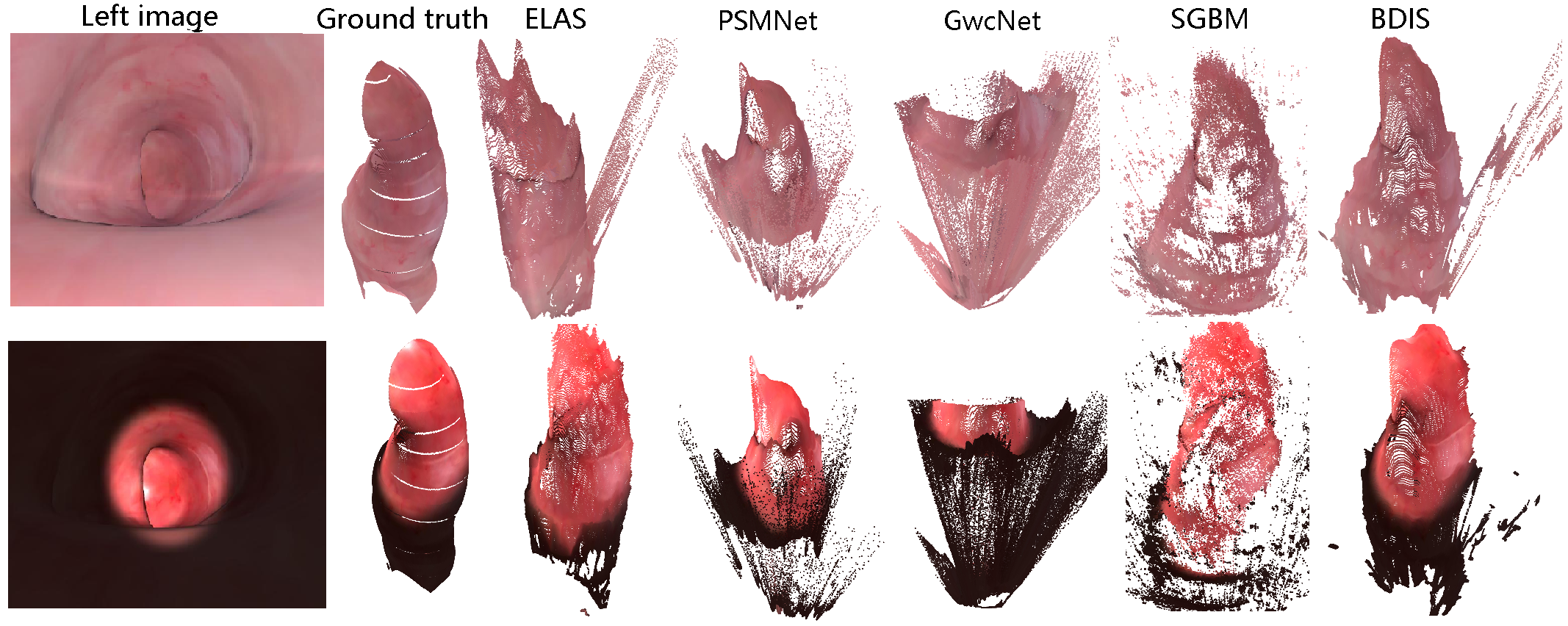}            
		\end{minipage}                
	}\\
	\caption{Sample reconstructions in Diffuse lighting and Lambertian reflectance scenarios.}
	\label{fig_exvivo_comparison}
\end{figure}
\begin{table*}[]
		\centering
		\caption{The results on the synthetic data with diffused light and Lambertian reflectance.}
		\begin{tabular}{p{2.12cm}<{\raggedright}|p{2.52cm}<{\centering}|p{1.12cm}<{\centering}|p{1.12cm}<{\centering}|p{1.12cm}<{\centering}|p{1.12cm}<{\centering}|p{1.12cm}<{\centering}|p{1.12cm}<{\centering}}
\hline
                                                                                      &                       & ELAS                            & SGBM   & DIS    & BDIS                            & GwcNet & PSMnet \\ \hline
\multirow{3}{*}{\begin{tabular}[c]{@{}l@{}}Diffuse\\ light\end{tabular}}              & Median error          & 0.178                           & 0.512  & 0.251  & \textbf{0.161} & 0.542  & 0.417  \\
                                                                                      & Mean error            & \textbf{0.220} & 1.113  & 0.753  & 0.320                           & 0.809  & 0.641  \\
                                                                                      & Valid pixels ($1000$) & 166.77                          & 103.92 & 288.41 & 208.44                          & 100.00 & 301.42 \\ \hline
\multirow{3}{*}{\begin{tabular}[c]{@{}l@{}}Non-Lambertian\\ reflectance\end{tabular}} & Median error          & 0.198                           & 0.710  & 0.376  & \textbf{0.163} & 0.271  & 0.731  \\
                                                                                      & Mean error            & \textbf{0.235} & 1.400  & 1.051  & 0.379                           & 0.662  & 1.027  \\
                                                                                      & Valid pixels ($1000$) & 81.50                           & 74.29  & 295.92 & 204.42                          & 106.38 & 301.46 \\ \hline
\end{tabular}
		\label{Table_exvivo_dataset}
	\end{table*}
	
	\subsection{Quantitative comparisons on the synthetic data set}

	BDIS was compared quantitatively with the baseline methods ELAS, SGBM, DIS, PSMNet, and GwcNet. The comparison between the prior-based DNN-based method and BDIS is for completeness only. The default setting of PSMNet and GwcNet were strictly followed. The pre-trained networks were adopted and finetuned with the labeled 300 (training) and 50 (validation) synthetic images for training and validation. Both were trained with Adam optimizer in 300 epochs.\par
	
	Table \ref{Table_exvivo_dataset} and Fig. \ref{fig_exvivo_comparison} show the comparisons on the synthetic data set, which are unaffected by distortion and inaccurate camera parameters. Considering the median error, BDIS is the best and has $9.55\%$ and $17.68\%$ higher accuracy than ELAS in diffuse lighting and non-Lambertian reflectance. The results indicate that BDIS is more advantageous in the scenario of non-Lambertian reflectance over ELAS, thus more robust in surgical scenarios. Results also show that BDIS cannot handle the edges well. Fig. \ref{fig_exvivo_comparison} and Table \ref{Table_exvivo_dataset} reveal the bad mean error comparison is attributed to the small group of far-out points on the dark regions/edges. The number of valid prediction suggest BDIS produces more predictions but suffers from inaccurate dark region predictions.\par 
	
	Readers may notice the bad performance of DNN, which contradicts the conclusion from \cite{allan2021stereo}. The reason is that the finetuning training process does not yield satisfying model parameters. The synthetic data set for transfer learning and the data used to pre-train the DNN are significantly different in terms of textures. Studies  \cite{chen2021s2r,song2021combining} indicate that the performance of the convolutional DNN is heavily dependent on the image texture, and efforts were devoted to bridging the domain gap \cite{chen2021s2r,zheng2018t2net}. The bad training process indicates its strong dependency on the training data set, which can be avoided using prior-free methods. Further tests will be conducted on labeled in-vivo data set.\par

	\subsection{Qualitative comparisons on the in-vivo dataset}
	\label{section_quantitative}
	
	We compared ELAS, BDIS, DIS, and SGBM on the in-vivo data sets. Since no ground truth is provided, DNN-based methods cannot be implemented. We aim to show that BDIS achieves similar accuracy as ELAS since near real-time ELAS is widely used in the community. Based on the scope-to-surface distance, the samples were categorized into five groups. Results show that BDIS achieves an average $0.4-1.66 mm$ (median error) and $0.65-2.32 mm$ (mean error) deviation from ELAS's results.\par

\begin{figure}[!h]
		\centering
		\subfloat{
			\begin{minipage}[]{1\textwidth}
				\centering
				\includegraphics[width=1\linewidth]{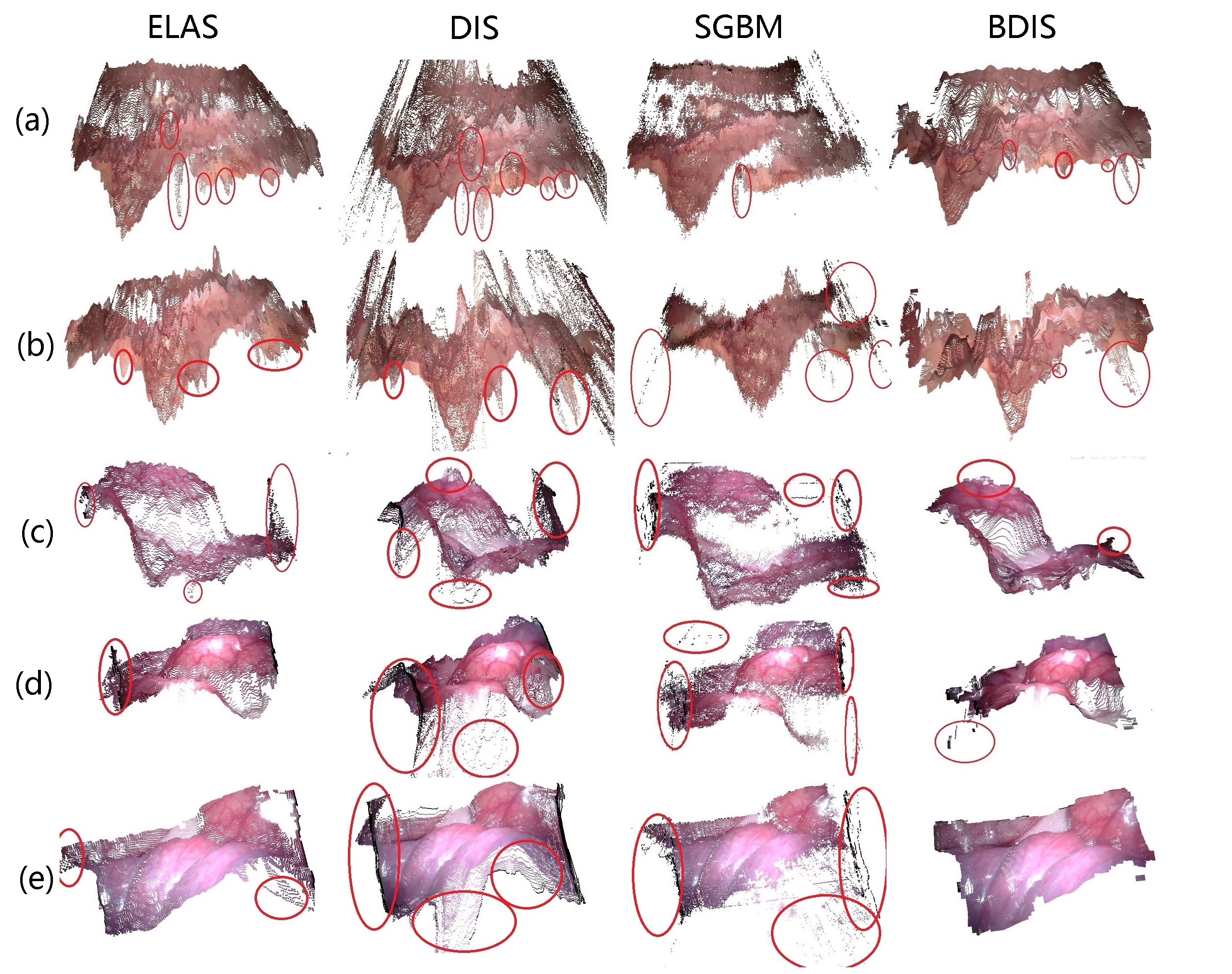}
			\end{minipage}
		}
		\caption{Sample recovered shapes of the 5 classes. Circles mark the regions with large error.}
		\label{fig_quantitative_compare}
	\end{figure}

	The invalid/dark/bright pixels lead to photometric inconsistency in the stereo matching process. Fig. \ref{fig_quantitative_compare} shows the qualitative comparisons of ELAS, DIS, SGBM, and BDIS on the relatively well-textured images. Generally, BDIS achieves similar performance as ELAS but better matches pixels at the image edge with fewer outliers. DIS and SGBM suffer from the wrong edges. Invalid pixels inevitably exist on the edges of the rectified image after the image undistortion. Thus, in the coarse-level patch disparity estimation, patches with more invalid pixels are more likely to fail in convergence or yield local minima (abnormal depth) due to insufficient information. The dubious predictions, however, substantially influence the prediction and the initialization of the disparity at the finer-scale patch optimization (as in \eqref{Eq_residual_fusion}). BDIS solves the problem by quantifying the posterior probability, discarding the patch that does not converge, and lowering the patches' probabilities with invalid pixels. Although the discarded patch does not help yield disparity, other patches compensate for the loss. If one pixel is not covered by any patch, we follow ELAS not to optimize the pixel.\par
	
	Another noticeable problem is the ambiguous local minima in the cost function, which penalizes the photometric inconsistency. Fig. \ref{fig_quantitative_compare} shows BDIS has fewer local minima than DIS and SGBM and is similar to ELAS. Fig. \ref{fig_quantitative_compare} (a-b) indicates that the BDIS addresses the patchs' probabilities with textured and alleviates the ambiguous disparity from the textureless surface. Fig. \ref{fig_quantitative_compare} (c-e) show that the ambiguities caused by the illumination have been greatly reduced. The quantitative results also provide evidence on its side. It should be emphasized that this work does not enforce any prior smoothness constraint in the optimization process. \par   

\begin{figure}[!h]
		\centering
		\subfloat{
			\begin{minipage}[]{0.95\textwidth}
				\centering
				\includegraphics[width=1\linewidth]{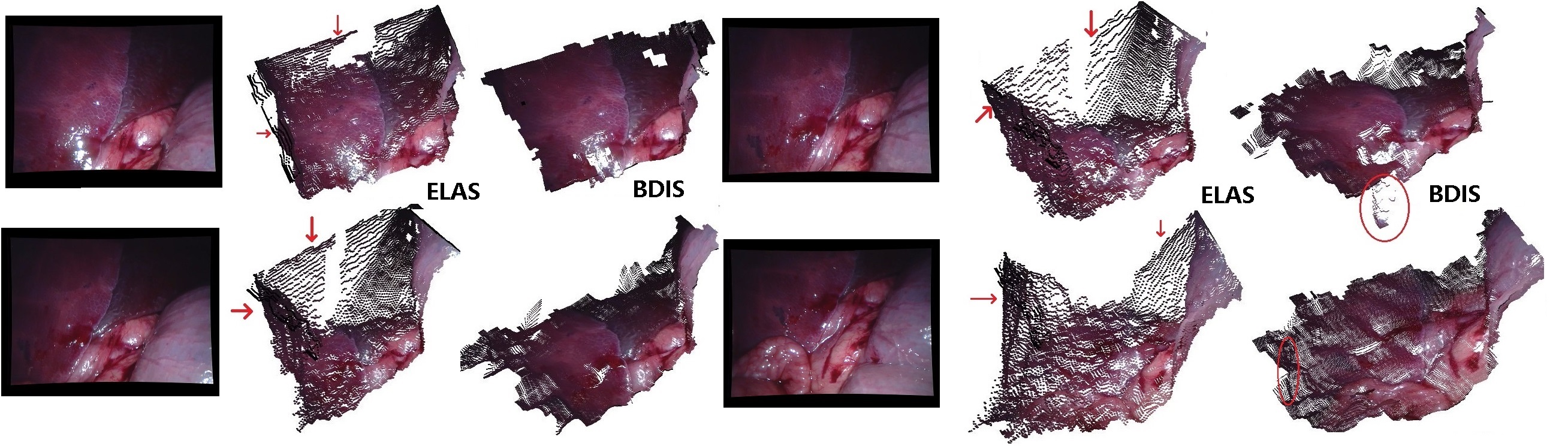}
			\end{minipage}
		}
		\caption{The qualitative comparisons on the heavy Lambertian reflectance and dark case.}
		\label{fig_quantitative_ELAS}
	\end{figure}

	We additionally tested BDIS and ELAS on the surfaces with serious non-Lambertian reflectance (Fig. \ref{fig_quantitative_ELAS}). The photometric consistency of this data deteriorates significantly. Fig. \ref{fig_quantitative_ELAS} shows that the center of the soft tissue is exposed to intense lighting while the marginal region is dark. Fig. \ref{fig_quantitative_ELAS} indicates that ELAS suffers from the ambiguity on the marginal dark regions while BDIS can ignore or estimate most dark pixels correctly. \par

	\subsection{Processing rate comparison}
	We compared the time consumption of ELAS, DIS, and BDIS on a single core of CPU (i5-9400). BDIS runs $10 Hz$ on $640\times480$ image and $25 Hz$ on $360\times288$ image while ELAS achieves $4 Hz$ and $11 Hz$. The two DNN methods GwcNet and PSMnet run $3 Hz$ and $5 Hz$ on GTX 1080ti. BDIS consumes double the time of DIS. The majority of the extra time of BDIS is devoted to patch-wise window traversing. Since the sampling window size is 5 in the experiment, 5 more times residual estimations are needed. In general, BDIS achieves similar/better performance over ELAS but runs 2 times faster. \par


	\section{Conclusion}
	
	We propose BDIS, the first CPU-level real-time stereo matching approach for CAS. BDIS inherits the fast performance of DIS while being more robust to textureless/dark surface and severe non-Lambertian reflectance. It achieves similar or better performance in accuracy as the near real-time method ELAS. A Bayesian approach and a spatial GMM are developed to describe the relative confidence of the pixel-wise disparity to achieve the performance. Experiments indicate that BDIS has fewer outliers than DIS and achieves a lower amount of outlier predictions than the near real-time ELAS. \par

	

	%
	%
	%
	\bibliographystyle{splncs04}
	\bibliography{bib/strings-full,bib/ieee-full,endoscope}
	%
	%
	%
	%
	%
\end{document}